\documentclass{article}

\usepackage{amsmath,amssymb,mylogic,isabelle,isabellesym}

\title{A formally verified proof of the \\prime number
  theorem\footnote{To appear in \emph{ACM Transcations on
      Computational Logic.} Work by the first author has been
    supported by NSF grant DMS-0401042.}}

\author{Jeremy Avigad, Kevin Donnelly, David Gray, and Paul Raff}

\date{January 19, 2006}

\begin{document}

\maketitle

\begin{abstract}
  The prime number theorem, established by Hadamard and de la Vall\'ee
  Poussin independently in 1896, asserts that the density of primes in
  the positive integers is asymptotic to $1 / \ln x$. Whereas their
  proofs made serious use of the methods of complex analysis,
  elementary proofs were provided by Selberg and Erd\"os in 1948.  We
  describe a formally verified version of Selberg's proof, obtained
  using the Isabelle proof assistant.
\end{abstract}

\section{Introduction}

For each positive integer $x$, let $\pi(x)$ denote the number of
primes less than or equal to $x$. The prime number theorem asserts
that the density of primes $\pi(x)/x$ in the positive integers is
asymptotic to $1 / \ln x$, i.e.~that $\lim_{x \to \infty} \pi(x) \ln x
/ x = 1$. This was conjectured by Gauss and Legendre around the turn
of the nineteenth century, and posed a challenge to the mathematical
community for almost a hundred years, until Hadamard and de la
Vall\'ee Poussin proved it independently in 1896.

On September 6, 2004, the first author of this article verified the
following statement, using the Isabelle proof assistant:
\begin{center}
\begin{isabellebody}
\isamarkupfalse%
{\isacharparenleft}{\isasymlambda}x{\isachardot}\ pi\ x\ {\isacharasterisk}\ ln\ {\isacharparenleft}real\ x{\isacharparenright}\ {\isacharslash}\ {\isacharparenleft}real\ x{\isacharparenright}{\isacharparenright}\ {\isacharminus}{\isacharminus}{\isacharminus}{\isacharminus}{\isachargreater}\ {\isadigit{1}}
\end{isabellebody}
\end{center}
The system thereby confirmed that the prime number theorem is a
consequence of the axioms of higher-order logic, together with an axiom
asserting the existence of an infinite set. 

One reason the formalization is interesting is simply that it is a
landmark, showing that today's proof assistants have achieved a level
of usability that makes it possible to formalize substantial theorems
of mathematics. Similar achievements in the past year include George
Gonthier's verification of the four color theorem using Coq, and
Thomas Hales's verification of the Jordan curve theorem using
HOL-light (see the introduction to Wiedijk \cite{wiedijk:05}). As
contemporary mathematical proofs become increasingly complex, the need
for formal verification becomes pressing. Formal verification can also
help guarantee correctness when, as is becoming increasingly common,
proofs rely on computations that are too long to check by hand.
Hales's ambitious Flyspeck project \cite{hales:unp}, which aims for a
fully verified form of his proof of the Kepler conjecture, is a
response to both of these concerns. Here, we will provide some
information as to the time and effort that went into our
formalization, which should help gauge the feasibility of such
verification efforts.

More interesting, of course, are the lessons that can be learned.
This, however, puts us on less certain terrain. Our efforts certainly
provide some indications as to how to improve libraries and systems
for verifying mathematics, but the data still need to be processed
and better understood. Here, therefore, we simply offer some initial
thoughts and observations.

The outline of this paper is as follows. In
Section~\ref{background:section}, we provide some background on the
prime number theorem and the Isabelle proof assistant. In
Section~\ref{overview:section}, we provide an overview of Selberg's
proof, our formalization, and the effort involved. In
Section~\ref{reflection:section}, we discuss some interesting aspects
of the formalization: the use of asymptotic reasoning; calculations
with real numbers; casts between natural numbers, integers, and real
numbers; combinatorial reasoning in number theory; and the use of
elementary methods. Finally, in Section~\ref{conclusions:section} we
offer some brief conclusions.

Our formalization of the prime number theorem was a collaborative
effort on the part of Avigad, Donnelly, Gray, and Raff, building, of
course, on the efforts of the entire Isabelle development team. This
article was, however, written by Avigad, so opinions and speculation
contained herein should be attributed to him.


\section{Background}
\label{background:section}

\subsection{The prime number theorem}

The statement of the prime number theorem was conjectured by both
Gauss and Legendre, on the basis of computation, around the turn of
the nineteenth century. In a pair of papers published in 1851 and
1852, Chebyshev made significant advances towards proving it. Note
that we can write
\[
\pi(x) = \sum_{p \leq x} 1,
\]
where $p$ ranges over the prime numbers. Contrary to our notation
above, $x$ is usually treated as a real variable, making $\pi$ a step
function on the reals. Chebyshev defined, in addition, the functions
\[
\theta(x) = \sum_{p \leq x} \ln p
\]
and
\[
\psi(x) = \sum_{p^a \leq x} \ln p = \sum_{n \leq x} \Lambda(n),
\]
where 
\[
\Lambda(n) =
\left\{ \begin{array}{ll}
  \ln p & \mbox{if $n = p^a$, for some $a \geq 1$} \\
  0 & \mbox{otherwise.}
\end{array}\right.
\]
The functions $\theta$ and $\psi$ are more sensitive to the presence
of primes than $\pi$, and have nicer analytic properties. Chebyshev
showed that the prime number theorem is equivalent to the assertion
$\lim_{x \to \infty} \theta(x) / x = 1$, as well as to the assertion
$\lim_{x \to \infty} \psi(x) / x = 1$. He also provided bounds
\[
B < \pi(x) \ln x / x < 6 B / 5
\]
for sufficiently large $x$, where 
\[
B = \ln 2 / 2 + \ln 3 / 3 + \ln 5 / 5 - \ln 30 / 30 > 0.92
\]
and $6 B / 5 < 1.11$. So, as $x$ approaches infinity, $\pi(x) \ln
x / x$, at worst, oscillates between these two values.

In a landmark work of 1859, Riemann introduced the complex-valued
function $\zeta$ into the study of number theory. It was not until
1894, however, that von Mangoldt provided an expression for $\psi$
that reduced the prime number theorem, essentially, to showing that
$\zeta$ has no roots with real part equal to 1. This last step was
achieved by Hadamard and de la Vall\'ee Poussin, independently, in
1896. The resulting proofs make strong use of the theory of complex
functions.  In 1921, Hardy expressed strong doubts as to whether a
proof of the theorem was possible which did not depend, fundamentally,
on these ideas. In 1948, however, Selberg and Erd\"os found elementary
proofs based on a ``symmetry formula'' due to Selberg. (The nature of
the interactions between Selberg and Erd\"os at the time and the
influence of ideas is a subtle one, and was the source of tensions
between the two for years to come.) Since the libraries we had to work
with had only a minimal theory of the complex numbers and a limited
real analysis library, we chose to formalize the Selberg proof.

There are a number of good introductions to analytic number theory
(for example, \cite{apostol:76,jameson:04}). Edwards's \emph{Riemann's
  zeta function} \cite{edwards:01} is an excellent source of both
historical and mathematical information. A number of textbooks present
Selberg's proof, including those by Nathanson \cite{nathanson:00},
Shapiro \cite{shapiro:83}, and Hardy and Wright
\cite{hardy:wright:79}. We followed Shapiro's excellent presentation
quite closely, though we made good use of Nathanson's book as well.

We also had help from another source. Cornaros and Dimitracopoulos
\cite{cornaros:dimitracopoulos:94} have shown that the prime number
theorem is provable in a weak fragment of arithmetic, by showing how
to formalize Selberg's proof (based on Shapiro's presentation) in that
fragment.\footnote{For issues relating to the formalization of
  mathematics, and number theory in particular, in weak theories of
  arithmetic, see Avigad \cite{avigad:03b}.}  Their concerns were
different from ours: by relying on a formalization of higher-order
logic, we were allowing ourselves a logically stronger theory; on the
other hand, Cornaros and Dimitracopoulos were concerned solely with
axiomatic provability and not ease of formalization. Their work was,
however, quite helpful in stripping the proof down to its bare
essentials.  Also, since our libraries did not have a good theory of
integration, we had to take some care to avoid the mild uses of
analysis in the textbook presentations. Cornaros and Dimitracopoulos's
work was again often helpful in that respect.

\subsection{Isabelle}

Isabelle \cite{isabelle} is a generic proof assistant developed under
the direction of Larry Paulson at Cambridge University and Tobias
Nipkow at TU Munich. The HOL instantiation \cite{nipkow:et:al:02}
provides a formal framework that is a conservative extension of
Church's simple type theory with an infinite type (from which the
natural numbers are constructed), extensionality, and the axiom of
choice. Specifically, HOL extends ordinary type theory with set types,
and a schema for polymorphic axiomatic type classes designed by
Nipkow and implemented by Marcus Wenzel \cite{wenzel:97}. It also
includes a definite description operator (``THE''), and an indefinite
description operator (``SOME'').\footnote{The extension by set types
  is mild, since they are easily interpretable in terms of predicate
  types $\sigma \to \fn{bool}$. Similarly, the definite description
  operator can be eliminated, at least in principle, using Russell's
  well-known interpretation. It is the indefinite description
  operator, essentially a version of Hilbert's epsilon operator, that
  gives rise to the axiom of choice.  Though we occasionally used the
  indefinite description operator for convenience, these uses could
  easily be replaced by the definition description operator, and it is
  likely that uses of the axiom of choice can be dispensed with in the
  libraries as well. In any event, it is a folklore result that
  G\"odel's methods transfer to higher-order logic to show that the
  axiom of choice is a conservative extension for a fragment the
  includes the prime number theorem.}

Isabelle offers good automated support, including a term simplifier,
an automated reasoner (which combines tableau search with rewriting),
and decision procedures for linear and Presburger arithmetic. It is an
LCF-style theorem prover, which is to say, correctness is guaranteed
by the use of a small number of constructors, in an underlying typed
programming language, to build proofs. Using the Proof General
interface \cite{proof:general}, one can construct proofs interactively
by repeatedly applying ``tactics'' that reduce a current subgoal to
simpler ones. But Isabelle also allows one to take advantage of a
higher-level proof language, called Isar, implemented by Wenzel
\cite{wenzel:02}. These two styles of interaction can, furthermore, be
combined within a proof. We found Isar to be extremely helpful in
structuring complex proofs, whereas we typically resorted to
tactic-application for filling in low-level inferences. Occasionally,
we also made mild use of Isabelle's support for locales
\cite{ballarin:05}. For more information on Isabelle, one should
consult the tutorial \cite{nipkow:et:al:02} and other online
documentation \cite{isabelle}.

Our formalization made use of the basic HOL library, as well as those
parts of the HOL-Complex library, developed primarily by Jacques
Fleuriot, that deal with the real numbers. Some of our earlier
definitions, lemmas, and theorems made their way into the 2004 release
of Isabelle, in which the formalization described here took
place. Some additional theorems in our basic libraries will be part of
the 2005 release. 


\section{Overview}
\label{overview:section}

\subsection{The Selberg proof}
\label{selberg:subsection}

The prime number theorem describes the asymptotic behavior of a
function from the natural numbers to the reals. Analytic number theory
works by extending the domain of such functions to the real numbers,
and then providing a toolbox for reasoning about such functions. One
is typically concerned with rough characterizations of a function's
rate of growth; thus $f = O(g)$ expresses the fact that
for some constant $C$, $| f(x) | \leq C | g(x) |$ for every $x$.
(Sometimes, when writing $f = O(g)$, one really means that the
inequality holds except for some initial values of $x$, where $g$ is
$0$ or one of the functions is undefined; or that the inequality holds
when $x$ is large enough.)

For example, all of the following identities can be obtained using
elementary calculus:
\begin{equation*}
\begin{split}
\ln (1 + 1 / n) & = 1 / n + O(1/n^2) \\ 
\sum_{n \leq x} 1 / n  & = \ln x + O(1) \\
\sum_{n \leq x} \ln n & = x \ln x - x + O(\ln x)\\
\sum_{n \leq x} \ln n / n & = \ln^2 x / 2 + O(1)
\end{split}
\end{equation*}
In all of these, $n$ ranges over positive integers. The last three
inequalities hold whether one takes $x$ to be an integer or a real
number greater than or equal to $1$. The second identity reflects the
fact that the integral of $1 / x$ is $\ln x$, and the third reflects
the fact that the integral of $\ln x$ is $x \ln x - x$. A list of
identities like these form one part of the requisite background to the
Selberg proof.

Some of Chebyshev's results form another part. Rate-of-growth
comparisons between $\theta$, $\psi$, and $\pi$ sufficient to show the
equivalence of the various statements of the prime number theorem can
be obtained by fairly direct calculations. Obtaining any of the upper
bounds equivalent to $\psi(x) = O(x)$ requires more work. A nice way
of doing this, using binomial coefficients, can be found in
Nathanson \cite{nathanson:00}.

Number theory depends crucially on having different ways of counting
things, and rudimentary combinatorial methods form a third
prerequisite to the Selberg proof. For example, consider the set of
(positive) divisors $d$ of a positive natural number $n$. Since the
function $d \mapsto n / d$ is a permutation of that set, we have the
following identity:
\[
\sum_{d | n} f(d) = \sum_{d | n} f(n / d).
\]
For a more complicated example, suppose $n$ is a positive integer, and
consider the set of pairs $d, d'$ of positive integers such that $d d'
\leq n$. There are two ways to enumerate these pairs: for each
value of $d$ between $1$ and $n$, we can enumerate all the values $d'$
such that $d' \leq n / d$; or for each product $c$ less than $n$, we
can enumerate all pairs $d, c / d$ whose product is $c$. Thus we have
\begin{equation}
\begin{split}
\sum_{d \leq n} \sum_{d' \leq n / d} f(d, d') & = \sum_{d d' \leq n}
f(d, d') \\ 
& = \sum_{c \leq n} \sum_{d | c} f(d, c/d).
\end{split}
\end{equation}
A similar argument yields
\begin{equation}
\label{counting:eq}
\begin{split}
\sum_{d | n} \sum_{d' | (n / d)} f(d, d') & = \sum_{d d' | n}
f(d, d') \\
& = \sum_{c | n} \sum_{d | c} f(d, c/d).
\end{split}
\end{equation}
Yet another important combinatorial identity is given by the partial
summation formula, which, in one formulation, is as follows: if $a
\leq b$, $F(n)
= \sum_{i = 1}^n f(i)$, and $G$ is any function, then
\begin{multline*}
\sum_{n = a}^b f(n + 1) G(n + 1) = F(b + 1) G(b + 1) - F(a) G(a + 1) -
\\
\sum_{n = a}^{b-1} F(n + 1) (G(n + 2) - G(n + 1)). 
\end{multline*}
This can be viewed as a discrete analogue of integration by parts, and
can be verified by induction.

An important use of (\ref{counting:eq}) occurs in the proof of the
M\"obius inversion formula, which we now describe. A positive natural
number $n$ is said to be \emph{square free} if no prime in its
factorization occurs with multiplicity greater than 1; in other words,
$n = p_1 p_2 \cdots p_s$ where the $p_i$'s are distinct primes (and
$s$ may be $0$). Euler's function $\mu$ is defined by
\[
\mu(n) = \left\{ 
  \begin{array}{ll}
  (-1)^s & \mbox{if $n$ is squarefree and $s$ is as above} \\
  0 & \mbox{otherwise.} \\
  \end{array}
\right.
\]
A remarkably useful fact regarding $\mu$ is that for $n > 0$, 
\begin{eqnarray}
\label{mu:eq}
\sum_{d | n} \mu(d) = \left\{
  \begin{array}{ll}
  1 & \mbox{if $n = 1$} \\
  0 & \mbox{otherwise.}
  \end{array}
\right.
\end{eqnarray}
To see this, define the \emph{radical} of a number $n$, denoted
$\fn{rad}(n)$, to be the greatest squarefree number dividing $n$. It
is not hard to see that if $n$ has prime factorization $p_1^{j_1}
p_2^{j_2} \cdots p_s^{j_s}$, then $\fn{rad}(n)$ is given by $p_1 p_2
\cdots p_s$. Then $\sum_{d | n} \mu(d) = \sum_{d | \fn{rad}(n)}
\mu(d)$, since divisors of $n$ that are not divisors of $\fn{rad}(n)$
are not squarefree and hence contribute $0$ to the sum. If $n = 1$,
equation (\ref{mu:eq}) is clear. Otherwise, write $\fn{rad}(n) = p_1
p_2 \cdots p_s$, write
\[
\sum_{d|\fn{rad}(n)} \mu(d) = \sum_{d | \fn{rad}(n), p_1 | d} \mu(d) + 
\sum_{d | \fn{rad}(n), p_1 \nmid d} \mu(d),
\]
and note that each term in the first sum is canceled by a
corresponding one in the second.

Now, suppose $g$ is any function from $\N$ to $\mathbb{R}$, and define $f$
by $f(n) = \sum_{d |n} g(d)$. The M\"obius inversion formula provides
a way of ``inverting'' the definition to obtain an expression for $g$
in terms of $f$.  Using (\ref{counting:eq}) for the third equality
below and (\ref{mu:eq}) for the last, we have, somewhat miraculously,
\begin{eqnarray*}
\sum_{d | n} \mu(d) f(n / d) & = &
  \sum_{d | n} \mu(d) \sum_{d' | (n / d)} g((n / d) / d') \\
& = & \sum_{d | n} \sum_{d' | (n / d)} \mu(d) g((n / d) / d') \\
& = & \sum_{c | n} \sum_{d | c} \mu(d) g(n / c) \\
& = & \sum_{c | n} g(n / c) \sum_{d | c} \mu(d) \\
& = & g(n),
\end{eqnarray*}
since the inner sum on the second-to-last line is $0$ except when $c$
is equal to $1$.

All the pieces just described come together to yield additional
identities involving sums, $\ln$, and $\mu$, as well as Mertens's
theorem:
\[
\sum_{n \leq x} \Lambda(n) / n = \ln x + O(1). 
\]
These, in turn, are used to derive Selberg's elegant ``symmetry
formula,'' which is the central component in the proof. One
formulation of the symmetry formula is as follows:
\[
\sum_{n \leq x} \Lambda (n) \ln n + \sum_{n \leq x} \sum_{d | n}
\Lambda(d) \Lambda(n / d) = 2 x \ln x + O(x).
\]
There are, however, many variants of this identity, involving
$\Lambda$, $\psi$, and $\theta$. These crop up in profusion because
one can always unpack definitions of the various functions, apply the
types of combinatorial manipulations described above, and use
identities and approximations to simplify expressions. 

What makes the Selberg symmetry formula so powerful is that there are
two terms in the sum on the left, each sensitive to the presence of
primes in different ways. The formula above implies there have to be
some primes --- to make left-hand side nonzero --- but there can't be
too many. Selberg's proof involves cleverly balancing the two terms
off each other, to show that in the long run, the density of the
primes has the appropriate asymptotic behavior.

Specifically, let $R(x) = \psi(x) - x$ denote the ``error term,'' and
note that by Chebyshev's equivalences the prime number theorem amounts
to the assertion $\lim_{x \to \infty} R(x) / x = 0$. With some
delicate calculation, one can use the symmetry formula to obtain a
bound on $|R(x)|$:
\begin{equation}
\label{r:eq}
|R(x)| \ln^2 x \leq 2 \sum_{n \leq x} | R(x / n)| \ln n + O(x \ln x).
\end{equation}
Now, suppose we have a bound $| R(x) | \leq a x$ for sufficiently large
$x$. Substituting this into the right side of (\ref{r:eq}) and using
an approximation for $\sum_{n \leq x} \ln n / n$ we get
\[
|R(x)| \leq a x + O(x / \ln x),
\]
which is not an improvement on the bound $|R(x)| \leq a x$ with which
we began. Selberg's method involves showing that in fact there are
always enough sufficiently large intervals on which one can obtain a
stronger bound on $R(x)$, so that for some positive constant $k$,
assuming we have a bound $|R(x)| \leq a x$ that valid for $x \geq
c_1$, we can obtain a $c_2$ and a better bound $|R(x)| \leq (a - k
a^3)$, valid for $x \geq c_2$. The constant $k$ depends on $a$, but
the same constant also works for any $a' < a$.

By Chebyshev's theorem, we know that there is a constant $a_1$ such
that $|R(x)| \leq a_1 x$ for every $x$. Choosing $k$ appropriate for
$a_1$ and then setting $a_{n+1} = a_n - k a_n^3$, we have that for
every $n$, there is a $c$ large enough so that $|R(x)|/x \leq a_n$ for
every $x \geq c$. But it is not hard to verify that the sequence $a_1,
a_2, \ldots$ approaches $0$, which implies that $R(x) / x$ approaches
$0$ as $x$ approaches infinity, as required.

\subsection{Our formalization}
\label{formalization:subsection}

All told, our number theory session, including the proof of the prime
number theorem and supporting libraries, constitutes 673 pages of
proof scripts, or roughly 30,000 lines.  This count includes about 65
pages of elementary number theory that we had at the outset, developed
by Larry Paulson and others; also about 50 pages devoted to a proof of
the law of quadratic reciprocity and properties of Euler's $\ph$
function, neither of which are used in the proof of the prime number
theorem.  The page count does not include the basic HOL library, or
properties of the real numbers that we obtained from the HOL-Complex
library.

The overview provided in the last section should provide a general
sense of the components that are needed for the formalization. To
start with, one needs good supporting libraries:
\begin{itemize}
\item a theory of the natural numbers and integers, including
  properties of primes and divisibility, and the fundamental theorem
  of arithmetic
\item a library for reasoning about finite sets, sums, and products
\item a library for the real numbers, including properties of $\ln$
\end{itemize}
The basic Isabelle libraries provided a good starting point, though we
had to augment these considerably as we went along. More specific
supporting libraries include:
\begin{itemize}
\item properties of the $\mu$ function, combinatorial identities, and
  the M\"obius inversion formula
\item a library for asymptotic ``big O'' calculations
\item a number of basic identities involving sums and $\ln$
\item Chebyshev's theorems
\end{itemize}
Finally, the specific components of the Selberg proof are:
\begin{itemize}
\item the Selberg symmetry formula
\item the inequality involving $R(n)$
\item a long calculation to show $R(n)$ approaches $0$
\end{itemize}

This general outline is clearly discernible in the list of theory
files, which can be viewed online \cite{avigad:isabelle}. Keep in
mind that the files described here have not been modified since the
original proof was completed, and many of the proofs were written
while various participants in the project were still learning how to
use Isabelle.  Since then, some of the basic libraries have been
revised and incorporated into Isabelle, but Avigad intends to revise
the number theory libraries substantially before cleaning up the rest
of the proof.

Once the basic libraries are in place, our formal proof follows
Shapiro's presentation quite closely, though for some parts we
followed Nathanson instead. A detailed description of
our proof would amount to little more than a step-by-step narrative
of (one of the various paths through) Selberg's proof, with page
correspondences in texts we followed. For example, one of our
formulations of the M\"obius inversion is as follows:

\medskip

\begin{isabellebody}
\isamarkupfalse%
\isacommand{lemma}\ mu{\isacharunderscore}inversion{\isacharunderscore}nat{\isadigit{1}}a{\isacharcolon}\ {\isachardoublequote}ALL\ n{\isachardot}\ {\isacharparenleft}{\isadigit{0}}\ {\isacharless}\ n\ {\isasymlongrightarrow}\isanewline
\ \ f\ n\ {\isacharequal}\ {\isacharparenleft}{\isasymSum}\ d\ {\isacharbar}\ d\ dvd\ n{\isachardot}\ g{\isacharparenleft}n\ div\ d{\isacharparenright}{\isacharparenright}{\isacharparenright}\ {\isasymLongrightarrow}\ {\isadigit{0}}\ {\isacharless}\ {\isacharparenleft}n{\isacharcolon}{\isacharcolon}nat{\isacharparenright}\ {\isasymLongrightarrow}\isanewline
\ \ g\ n\ {\isacharequal}\ {\isacharparenleft}{\isasymSum}\ d\ {\isacharbar}\ d\ dvd\ n{\isachardot}\ of{\isacharunderscore}int{\isacharparenleft}mu{\isacharparenleft}int{\isacharparenleft}d{\isacharparenright}{\isacharparenright}{\isacharparenright}\ {\isacharasterisk}\ f\ {\isacharparenleft}n\ div\ d{\isacharparenright}{\isacharparenright}{\isachardoublequote}
\end{isabellebody}

\medskip

\noindent This appears on page 64 of Shapiro's book, and on page 218
of Nathanson's book. We formalized a version of the fourth identity
listed in Section~\ref{formalization:subsection} as follows:

\medskip

\begin{isabellebody}
\isamarkupfalse%
\isacommand{lemma}\ identity{\isacharunderscore}four{\isacharunderscore}real{\isacharunderscore}b{\isacharcolon}\ {\isachardoublequote}{\isacharparenleft}{\isasymlambda}x{\isachardot}\ {\isasymSum}i{\isacharequal}{\isadigit{1}}{\isachardot}{\isachardot}natfloor{\isacharparenleft}abs\ x{\isacharparenright}{\isachardot}\isanewline
\ \ \ \ ln\ {\isacharparenleft}real\ i{\isacharparenright}\ {\isacharslash}\ {\isacharparenleft}real\ i{\isacharparenright}{\isacharparenright}\ {\isacharequal}o\ \isanewline
\ \ \ \ \ \ {\isacharparenleft}{\isasymlambda}x{\isachardot}\ ln{\isacharparenleft}abs\ x\ {\isacharplus}\ {\isadigit{1}}{\isacharparenright}{\isacharcircum}{\isadigit{2}}\ {\isacharslash}\ {\isadigit{2}}{\isacharparenright}\ {\isacharplus}o\ O{\isacharparenleft}{\isasymlambda}x{\isachardot}\ {\isadigit{1}}{\isacharparenright}{\isachardoublequote}
\end{isabellebody}

\medskip

\noindent In fact, stronger assertions can be found on page 93 of Shapiro's
book, and on page 209 of Nathanson's book. Here is one of our
formulations of the Selberg symmetry principle:

\medskip

\begin{isabellebody}
\isamarkupfalse%
\isacommand{lemma}\ Selberg{\isadigit{3}}{\isacharcolon}\ {\isachardoublequote}{\isacharparenleft}{\isasymlambda}x{\isachardot}\ {\isasymSum}n\ {\isacharequal}\ {\isadigit{1}}{\isachardot}{\isachardot}natfloor\ {\isacharparenleft}abs\ x{\isacharparenright}\ {\isacharplus}\ {\isadigit{1}}{\isachardot}\isanewline
\ \ \ \ Lambda\ n\ {\isacharasterisk}\ ln\ {\isacharparenleft}real\
n{\isacharparenright}{\isacharparenright}\ {\isacharplus}\ {\isacharparenleft}{\isasymlambda}x{\isachardot}\ {\isasymSum}n{\isacharequal}{\isadigit{1}}{\isachardot}{\isachardot}natfloor\ {\isacharparenleft}abs\ x{\isacharparenright}\ {\isacharplus}\ {\isadigit{1}}{\isachardot}\ \isanewline
\ \ \ \ \ {\isacharparenleft}{\isasymSum}\ u\ {\isacharbar}\ u\ dvd\ n{\isachardot}\ Lambda\ u\ {\isacharasterisk}\ Lambda\ {\isacharparenleft}n\ div\ u{\isacharparenright}{\isacharparenright}{\isacharparenright}\isanewline
\ \ \ \ \ \ {\isacharequal}o\ {\isacharparenleft}{\isasymlambda}x{\isachardot}\ {\isadigit{2}}\ {\isacharasterisk}\ {\isacharparenleft}abs\ x\ {\isacharplus}\ {\isadigit{1}}{\isacharparenright}\ {\isacharasterisk}\ ln\ {\isacharparenleft}abs\ x\ {\isacharplus}\ {\isadigit{1}}{\isacharparenright}{\isacharparenright}\ {\isacharplus}o\ O{\isacharparenleft}{\isasymlambda}x{\isachardot}\ abs\ x\ {\isacharplus}\ {\isadigit{1}}{\isacharparenright}{\isachardoublequote}
\end{isabellebody}

\medskip

\noindent This is given on page 419 of Shapiro's book, and on page 293 of
Nathanson's book. The error estimate given in the previous section,
taken from 431 of Shapiro's book, takes the following form:

\medskip

\begin{isabellebody}
\isamarkupfalse%
\isacommand{lemma}\ error{\isadigit{7}}{\isacharcolon}\ {\isachardoublequote}{\isacharparenleft}{\isasymlambda}x{\isachardot}\ abs\ {\isacharparenleft}R\ {\isacharparenleft}abs\ x\ {\isacharplus}\ {\isadigit{1}}{\isacharparenright}{\isacharparenright}\ {\isacharasterisk}\ ln\ {\isacharparenleft}abs\ x\ {\isacharplus}\ {\isadigit{1}}{\isacharparenright}\ {\isacharcircum}\ {\isadigit{2}}{\isacharparenright}\ {\isacharless}o\isanewline
\ \ {\isacharparenleft}{\isasymlambda}x{\isachardot}\ {\isadigit{2}}\ {\isacharasterisk}\ {\isacharparenleft}{\isasymSum}\ n\ {\isacharequal}\ {\isadigit{1}}{\isachardot}{\isachardot}natfloor\ {\isacharparenleft}abs\ x{\isacharparenright}\ {\isacharplus}\ {\isadigit{1}}{\isachardot}\isanewline
\ \ \ \ \ \ abs\ {\isacharparenleft}R\ {\isacharparenleft}{\isacharparenleft}abs\ x\ {\isacharplus}\ {\isadigit{1}}{\isacharparenright}\ {\isacharslash}\ real\ n{\isacharparenright}{\isacharparenright}\ {\isacharasterisk}\ ln\ {\isacharparenleft}real\ n{\isacharparenright}{\isacharparenright}{\isacharparenright}\ {\isacharequal}o\ \isanewline
\ \ \ \ O{\isacharparenleft}{\isasymlambda}x{\isachardot}\ {\isacharparenleft}abs\ x\ {\isacharplus}\ {\isadigit{1}}{\isacharparenright}\ {\isacharasterisk}\ {\isacharparenleft}{\isadigit{1}}\ {\isacharplus}\ ln\ {\isacharparenleft}abs\ x\ {\isacharplus}\ {\isadigit{1}}{\isacharparenright}{\isacharparenright}{\isacharparenright}{\isachardoublequote}
\end{isabellebody}

\medskip

\noindent We will have more to say, below, about handling of 
asymptotic notation, the type casts, and the various occurrences of
$\fn{abs}$ and $+1$ that make the formal presentation differ from
ordinary mathematical notation. But aside from calling attention to
differences like these, a more detailed outline would not be very
interesting.

There are additional reasons that it does not pay to describe the
formal proofs in great detail. For one thing, they are not
particularly nice: our efforts were designed to get us to the prime
number theorem as quickly as possible, and so the proofs could use a
good deal of cleaning and polishing. Second, and more to the point, we
know that our formalization is not optimal. It hardly makes sense for
us to describe exactly how we went about proving the M\"obius
inversion formula, for example, until we are convinced that we have
done it right; that is, until we are convinced the we have made the
supporting libraries as generally useful as possible, and configured
the automated tools in such a way to make the formalization as smooth
as possible. We therefore intend to invest more time in improving the
various parts of the formalization and report on these when it is
clear what we have learned from the efforts.

In the meanwhile, we will devote the rest of this report to conveying
two types of information. First, to help gauge the usability of the
current technology, we will try to provide a sense of the amount of
time required to seeing the project through to its completion. Second,
we will provide some initial reflections on the project, and on the
strengths and weaknesses of contemporary proof assistants. In
particular, we will discuss what we take to be some of the novel
aspects of the formalization, and indicate where we believe better
automated support would have been especially helpful.

\subsection{The effort involved}

As we have noted in the introduction, one of the most interesting
features of our formalization of the prime number theorem is simply
its existence, which shows that current technology makes it possible
to treat a proof of this complexity. The question naturally arises as
to how long the formalization took.

This is a question that it hard to answer with any precision.  Avigad
first decided to undertake the project in March of 2003, having
learned how to use Isabelle and proved Gauss's law of quadratic
reciprocity with Gray and Adam Kramer the preceding summer and fall.
But this was a side project for everyone involved, and time associated
it includes time spent learning to use Isabelle, time spent learning
the requisite number theory, and so on. Gray developed a substantial
part of the number theory library, including basic facts about primes
and multiplicity, the $\mu$ function, and the identity
(\ref{counting:eq}), working a few hours per week in the summer of
2003, before his thesis work in ethics took over. Donnelly and Avigad
developed the library to support big O calculations
\cite{avigad:donnelly:04} while Donnelly worked half-time during the
summer of 2003, just after he completed his junior year at Carnegie
Mellon. During that summer, and working part time the following year,
Donnelly also derived some of the basic identities involving $\ln$.
Raff started working on the project in the 2003-2004 academic year,
but most of his contributions came working roughly half-time in the
summer of 2004, just after he obtained his undergraduate degree.
During that time, he proved Chebyshev's theorem to the effect that
$\psi(x) = O(x)$, and also did most of the work needed to prove the
equivalence of statements of the prime number theorem in terms of the
functions $\pi$, $\theta$, and $\psi$. Though Avigad's involvement was
more constant, he rarely put in more than a few hours per week before
the summer of 2004, and set the project aside for long stretches of
time. The bulk of his proof scripts were written during the summer of
2004, when he worked roughly half-time on the project from the middle
of June to the end of August.

Some specific benchmarks may be more informative. Proving most of the
inversion theorems we needed, starting from (\ref{counting:eq}) and
the relevant properties of $\mu$, took Avigad about a day. (For a
``day'' read eight hours of dedicated formalization.  Though he could
put in work-days like that for small stretches, in some of the
estimates below, the work was spread out over longer periods of time.)
Proving the first version of the Selberg symmetry formula using the
requisite identities took another day. Along the way, he was often
sidetracked by the need to prove elementary facts about things like
primes and divisibility, or the floor function on the real numbers.
This process stabilized, however, and towards the end he found that he
could formalize about a page of Shapiro's text per day. Thus, the
derivation of the error estimate described above, taken from pages
428--431 in Shapiro's book, took about three-and-a-half days to
formalize; and the remainder of the proof, corresponding to 432--437
in Shapiro's book, took about five days.

In many cases, the increase in length is dramatic: the
three-and-a-half pages of text associated with the proof of the error
estimate translate to about 1,600 lines, or 37 pages, of proof
scripts, and the five pages of text associated with the final part of
the proof translate to about 4,000 lines, or 89 pages, of proof
scripts. These ratios are abnormally high, however, for reasons
discussed in Section~\ref{calculation:subsection}. The five-line
derivation of the M\"obius inversion formula in
Section~\ref{selberg:subsection} translates to about 40 lines, and the
proof of the form of the Selberg symmetry formula discussed there,
carried out in about two-and-a-half pages in Shapiro's book, takes up
about 600 lines, or 13 pages. These ratios are more typical.

We suspect that over the coming years both the time it takes to carry
out such formalizations, as well as the lengths of the formal proof
scripts, will drop significantly. Much of the effort involved in the
project was spent on the following:
\begin{itemize}
\item Defining fundamental concepts and gathering basic libraries of
  easy facts.
\item Proving trivial lemmas and spelling out ``straightforward''
  inferences.
\item Finding the right lemmas and theorems to apply.
\item Entering long formulas and expressions correctly, and adapting
  ordinary mathematical notation to the formal notation in Isabelle.
\end{itemize}
Gradually, all these requirements will be ameliorated, as better
libraries, automated tools, and interfaces are developed. On a
personal note, we are entirely convinced that, although there is a
long road ahead, formal verification of mathematics will inevitably
become commonplace. Getting to that point will require both
theoretical and practical ingenuity, but we do not see any conceptual
hurdles.\footnote{For further speculation along these lines, see the
  preliminary notes \cite{avigad:unp:pnt}.}


\section{Thoughts on the formalization}
\label{reflection:section}

In this section, we will discuss features of the formalization that we
feel are worthy of discussion, either because they represent novel and
successful solutions to general problems, or (more commonly) because
they indicate aspects of formal mathematical verification where better
support is possible.

\subsection{Asymptotics}

One of our earliest tasks in the formalization was to develop a
library to support the requisite calculations with big O expressions.
To that end, we gave the expression $f = O(g)$ the strict reading $\ex
C \fa x (|f(x)| \leq C |g(x)|)$, and followed the common practice of
taking $O(g)$ to be the set of all functions with the requisite rate
of growth, i.e.~
\[
O(g) = \{ f \st \ex C \fa x (|f(x)| \leq C |g(x)|) \}.
\]
We then read the ``equality'' in $f = O(g)$ as the element-of
relation, $\in$. 

Note that these expressions make sense for any function type for which
the codomain is an ordered ring. Isabelle's axiomatic type classes
made it possible to develop the library fully generally. We were able
to lift operations like addition and multiplication to such types,
defining $f + g$ to denote the pointwise sum, $\lambda x. (f(x) +
g(x))$. Similarly, given a set $B$ of elements of a type that supports
addition, we defined
\[
a +_o B = \{ c \st \ex {b \in B} (c = a + b) \}.
\]
We also defined $a =_o B$ to be alternative input syntax for $a \in
B$.  This gave expressions like $f =_o g +_o O(h)$ the intended
meaning. In mathematical texts, convention dictates that in an
expression like $x^2 + 3 x = x^2 + O(x)$, the terms are to be
interpreted as functions of $x$; in Isabelle we had to use lambda
notation to make this explicit. Thus, the expression above would be
entered
\begin{center}
{\isastyle {\isacharparenleft}{\isasymlambda}x{\isachardot}\
  x{\isacharcircum}{\isadigit{2}}\ {\isacharplus}\ {\isadigit{3}}\
  {\isacharasterisk}\ x{\isacharparenright}\ {\isacharequal}o\
  {\isacharparenleft}{\isasymlambda}x{\isachardot}\
  x{\isacharcircum}{\isadigit{2}}{\isacharparenright}\ {\isacharplus}o\
  O{\isacharparenleft}{\isasymlambda}x{\isachardot}\
  x{\isacharparenright}}
\end{center}
This should help the reader make sense of the formalizations presented
in Section~\ref{formalization:subsection}.

An early version of our big O library was presented at IJCAR
\cite{avigad:donnelly:04}. That version is nonetheless fairly close to
the version used in the proof of the prime number theorem described
here, as well as a version that is scheduled for the 2005 release of
Isabelle.\footnote{Improvements in the more recent versions include
  better and more general theorems involving summations, theorems to
  handle composition of big O equations, and support for reasoning
  about asymptotic inequalities. Also, in the most recent version, we
  have dispensed with expressions of the form $O(S)$, where $S$ is a
  set of functions. It seems that uses of these are easily eliminable,
  and having $O$ notation for both functions and sets of functions led
  to annoying type ambiguities.}

There is one feature of our library that seems to be less than
optimal, and resulted in a good deal of tedium. With our definition, a
statement like $\lam {x.} x + 1 = O(\lam {x.} x^2)$ is false when the
variables range over the natural numbers, since $x^2$ is equal to $0$
when $x$ is $0$. Often one wants to restrict one's attention to
strictly positive natural numbers, or nonnegative real numbers. There
are four ways one can do this:
\begin{itemize}
\item Define new types for the strictly positive natural numbers, or
  nonnegative real numbers, and state the identities for those types.
\item Formalize the notion ``$f = O(g)$ on $S$.''
\item Formalize the notion ``$f = O(g)$ eventually.''
\item Replace $x$ by $x + 1$ in the first case, and by $|x|$ in the
  second case, to make the identities correct. For example, ``$f(|x|)
  = O(|x|^3)$'' expresses that $f(x) = O(x^3)$ on the nonnegative
  reals. Various similar tinkerings are effective; for example, the
  relationship intended in the example above is probably best
  expressed as $\lam {x.} x + 1 = O(\lam{x.} x^2 + 1)$.
\end{itemize}
These various options are discussed in the IJCAR paper
\cite{avigad:donnelly:04}, and all come at a cost. For example, the
first requires annoying casts, say, between positive natural numbers,
and natural numbers. The second requires carrying around a set $S$ in
every formula, and both the second and third require additional work
when composing expressions or reasoning about sums (roughly, one has
to make sure that the range of a function lies in the domain where an
asymptotic estimate is valid).

In our formalization, we chose the fourth route, which explains the
numerous occurrences of $+1$ and $\fn{abs}$ in the statements in
Section~\ref{formalization:subsection}. This often made some of the
more complex calculations painfully tedious, forcing us, for example,
the following ``helper'' lemma in Selberg:

\medskip

\begin{isabellebody}
\isamarkupfalse%
\isacommand{lemma}\ aux{\isacharcolon}\ {\isachardoublequote}{\isadigit{1}}\ {\isacharless}{\isacharequal}\ z\ {\isasymLongrightarrow}\ natfloor{\isacharparenleft}abs{\isacharparenleft}z\ {\isacharminus}\ {\isadigit{1}}{\isacharparenright}{\isacharparenright}\ {\isacharplus}\ {\isadigit{1}}\ {\isacharequal}\ natfloor\ z{\isachardoublequote}
\end{isabellebody}

\medskip

\noindent On the general principle that formalization goes most
smoothly when the formalization is as close as possible to the
informal text, it is probably worth extending the library in the ways
described above. We do not have a good sense, however, as to how much
this would have simplified our task.

Donnelly and Avigad have designed a decision procedure for entailments
between linear big O equations, and have obtained a prototype
implementation (though we have not incorporated it into the Isabelle
framework). This would eliminate the need for helper lemmas like the
following:

\medskip 

\begin{isabellebody}
\isamarkupfalse%
\isacommand{lemma}\ aux{\isadigit{5}}{\isacharcolon}\ {\isachardoublequote}f\ {\isacharplus}\ g\ {\isacharequal}o\ h\ {\isacharplus}o\ O{\isacharparenleft}k{\isacharcolon}{\isacharcolon}{\isacharprime}a{\isacharequal}{\isachargreater}{\isacharparenleft}{\isacharprime}b{\isacharcolon}{\isacharcolon}ordered{\isacharunderscore}ring{\isacharparenright}{\isacharparenright}\ {\isasymLongrightarrow}\isanewline
\ \ g\ {\isacharplus}\ l\ {\isacharequal}o\ h\ {\isacharplus}o\ O{\isacharparenleft}k{\isacharparenright}\ {\isasymLongrightarrow}\ f\ {\isacharequal}o\ l\ {\isacharplus}o\ O{\isacharparenleft}k{\isacharparenright}{\isachardoublequote}
\end{isabellebody}

\medskip

\noindent We believe calculations going beyond the linear fragment
would also benefit from a better handling of monotonicity, just as is
needed to support ordinary calculations with inequalities,
as described in the next section.

\subsection{Calculations with real numbers}
\label{calculation:subsection}

One salient feature of the Selberg proof is the amount of calculation
involved. The dramatic increase in the length of the formalization of
the final part of the proof (5 pages in Shapiro, compared to 89 or so
in the formal version) is directly attributable to the need to spell
out calculations involving field operations, logarithms and
exponentiation, the greatest and least integer functions (``ceiling''
and ``floor''), and so on. The textbook calculations themselves were
complex; but then each textbook inference had to be expanded, by hand,
to what was often a long sequence of entirely straightforward
inferences.

Of course, Isabelle does provide some automated support. For example,
the simplifier employs a form of ordered rewriting for operations,
like addition and multiplication, that are associative and
commutative. This puts terms involving these operations into canonical
normal forms, thereby making it easy to verify equality of terms that
differ up to such rewriting. More complex equalities can similarly be
obtained by simplifying with appropriate rewrite rules, such as
various forms of distributivity in a ring or identities for logarithms
and exponents.

Much of the work in the final stages of the proof, however, involved
verifying \emph{inequalities} between expressions. Isabelle's linear
arithmetic package is complete for reasoning about inequalities
between linear expressions in the integers and reals, i.e.~validities
that depend only on the linear fragment of these theories. But,
many of the calculations went just beyond that,
at which point we were stuck manipulating expressions by hand and
applying low-level inferences.

As a simple example, part of one of the long proofs in
PrimeNumberTheorem required verifying that
\[
(1 + \frac{\varepsilon}{3(C + 3)}) \cdot n < K x
\]
using the following hypotheses:
\[
\begin{split}
& n \leq (K / 2) x \\
& 0 < C \\
& 0 < \varepsilon < 1
\end{split}
\]
The conclusion is easily obtained by noting that $1 +
\frac{\varepsilon}{3(C + 3)}$ is strictly less than $2$, and so the
product with $n$ is strictly less than $2 (K / 2) x = K x$.
But spelling out the details requires, for one thing, invoking the
relevant monotonicity rules for addition, multiplication, and
division. The last two, in turn, require verifying that the relevant
terms are positive. Furthermore, getting the calculation to go through
can require explicitly specifying terms like $2 (K / 2) x$ (which can
be simplified to $K x$), or, in other contexts, using rules like
associativity or commutativity to manipulate terms into the the forms
required by the rules.

The file PrimeNumberTheorem consists of a litany of such calculations.
This required us to have names like ``mult-left-mono''
``add-pos-nonneg,'' ``order-le-less-trans,'' ``exp-less-cancel-iff,''
``pos-divide-le-eq'' at our fingertips, or to search for them when
they were needed.  Furthermore, sign calculations had a way of coming
back to haunt us.  For example, verifying an inequality like $1 / (1 +
st) < 1 / (1 + su)$ might require showing that the denominators are
positive, which, in turns, might require verifying that $s$, $t$, and
$u$ are nonnegative; but then showing $s t > s u$ may again require
verifying that $s$ is positive. Since $s$ can be carried along in a
chain of inequalities, such queries for sign information can keep
coming back. Isar made it easy to break out such facts, name them, and
reuse them as needed. But since we were usually working in a context
where obtaining the sign information was entirely straightforward,
these concerns always felt like an annoying distraction from the
interesting and truly difficult parts of the calculations.

In short, inferences like the ones we have just described are commonly
treated as ``obvious'' in ordinary mathematical texts, and it would be
nice if mechanized proof assistants could recognize them as such.
Decision procedures that are stronger than linear arithmetic are
available; for example, a proof-producing decision procedure for
real-closed fields has recently been implemented in HOL-light
\cite{mclaughlin:harrison:05}. But for calculations like the one
above, computing sequences of partial derivatives, as decision
procedures for the real closed fields are required to do, is arguably
unnecessary and inefficient. Furthermore, decision procedures for real
closed fields cannot be extended, say, to handle exponentiation and
logarithms; and adding a generic monotone function, or trigonometric
functions, or the floor function, renders the full theory undecidable.

Thus, in contexts similar to ours, we expect that principled heuristic
procedures will be most effective. Roughly, one simply needs to chain
backwards through the obvious rules in a sensible way. There are
stumbling blocks, however. For one thing, excessive case splits can
lead to exponential blowup; e.g.~one can show $s t > 0$ by
showing that $s$ and $t$ are either both strictly positive or strictly
negative. Other inferences are similarly nondeterministic: one can
show $r + s + t > 0$ by showing that two of the terms are nonnegative
and the third is strictly positive, and one can show $r + s < t + u +
v + w$, say, by showing $r < u$, $s \leq t + v$, and $0 \leq w$.

As far as case splits are concerned, we suspect that they are rarely
needed to establish ``obvious'' facts; for example, in
straightforward calculations, the necessary sign information is
typically available. As far as the second
sort of nondeterminism is concerned, notice that the procedures for
linear arithmetic are effective in drawing the requisite conclusions
from available hypotheses; this is a reflection that of the fact that
the theory of the real numbers with addition (and, say, multiplication
by rational constants) is decidable. 

The analogous theory of the reals with multiplication is also
decidable. To see this, observe that the structure consisting of the
strictly positive real numbers with multiplication is isomorphic to
the structure of the real numbers with addition, and so the usual
procedures for linear arithmetic carry over. More generally, by
introducing case splits on the signs of the basic terms, one can
reduce the multiplicative fragment of the reals to the previous case.

In short, when the signs of the relevant terms are known, there are
straightforward and effective methods of deriving inequalities in the
additive and multiplicative fragments. This suggests that what is
really needed is a principled method of amalgamating such ``local''
procedures, together with, say, procedures that make use of
monotonicity and sign properties of logarithms and exponentiation. The
well-known Nelson-Oppen procedure provides a method of amalgamating
decision procedures for disjoint theories that share only the equality
symbol in their common language; but these methods fail for theories
that share an inequality symbol when one adds, say, rational constants
to the language, which is necessary to render such combinations
nontrivial. We believe that there are principled ways, however, of
extending the Nelson-Oppen framework to obtain useful heuristic
procedures. This possibility is explored in Avigad and Friedman
\cite{avigad:friedman:unp}.

\subsection{Casting between domains}
\label{casting:subsection}

In our formalization, we found that the most natural way to establish
basic properties of the functions $\theta$, $\psi$, and $\pi$, as well
as Chebyshev's theorems, was to treat them as functions from the
natural numbers to the reals, rather them as functions from the reals
to the reals. Either way, however, it is clear that the relevant
proofs have to use the embedding of the natural numbers into the reals
in an essential way. Since the $\mu$ function takes positive and
negative values, we were also forced to deal with integers as soon
as $\mu$ came into play. In short, our proof of the prime number
theorem inevitably involved combining reasoning about the natural
numbers, integers, and real numbers effectively; and this, in turn,
involved frequent casting between the various domains.

We tended to address such needs as they arose, in an ad-hoc way.  For
example, the version of the fundamental theorem of arithmetic that we
inherited from prior Isabelle distributions asserts that every
positive natural number can be written uniquely as the product of an
increasing list of primes. Developing properties of the radical
function required being able to express the unique factorization
theorem in the more natural form that every positive number is the
product of the primes that divide it, raised to the appropriate
multiplicity; i.e.~the fact that for every $n > 0$, 
\[
n = \prod_{p | n} p^{\fn{mult}_p(n)}, 
\]
where $\fn{mult}_p(n)$ denotes the multiplicity of $p$ in $n$. We also
needed, at our disposal, things like the fact that $n$ divides $m$ if
and only if for every prime number $p$, the multiplicity of $p$ in $n$
is less than or equal to the multiplicity of $p$ in $m$. Thus, early
on, we faced the dual tasks of translating the unique factorization
theorem from a statement about positive natural numbers to positive
integers, and developing a good theory of multiplicity in that
setting. Later, when proving Chebyshev's theorems, we found that we
needed to recast some of the facts about multiplicity to statements
about natural numbers.

We faced similar headaches when we began serious calculations
involving natural numbers and the reals. In particular, as we
proceeded we were forced to develop a substantial theory of the floor
and ceiling functions, including a theory of their behavior vis-a-vis
the various field operations. In calculations, expressions sometimes
involved objects of all three types, and we often had to explicitly
transport operations in or out of casts in order to apply a relevant
lemma.

When one extends a domain like the natural numbers to the integers, or
the integers to the real numbers, some operations are simply extended.
For example, properties of addition and multiplication of natural
numbers carry all the way through to the reals. On the other hand, one
has new operations, like subtraction on the integers and division in
the real numbers, that are mirrored imperfectly in the smaller
domains. For example, subtraction on the integers extends truncated
subtraction $x \tsub y$ on the natural numbers only when $x \geq y$,
and division in the reals extends the function $x \mathop{\fn{div}} y$
on the integers or natural numbers only when $y$ divides $x$. Finally,
there are facts that depend on the choice of a left inverse to the
embedding: for example, if $n$ is an integer, $x$ is a real number,
$\fn{real}$ is the embedding of the integers into the reals, and
$\lfloor \cdot \rfloor$ denotes the floor function from the reals to
the integers, we have
\[
(n \leq \lfloor x \rfloor) \equiv (\fn{real}(n) \leq x).
\]
This is an example of what mathematicians call a
Galois correspondence, and category theorists call an adjunction,
between the integers and the real numbers with the ordering relation.

Our formalization of the prime number theorem involved a good deal of
manipulation of expressions, by hand, using the three types of facts
just described. Many of these inferences should be handled
automatically. After all, such issues are transparent in mathematical
texts; we carry out the necessary inferences smoothly and
unconsciously whenever we read an ordinary proof. The guiding
principle should be that anything that is transparent to us can be
made transparent to a mechanized proof assistant: we simply need to
reflect on \emph{why} we are effectively able to combine domains in
ordinary mathematical reasoning, and codify that knowledge
appropriately.

\subsection{Combinatorial reasoning with sums}

As described in Section~\ref{formalization:subsection}, formalizing
the prime number theorem involved a good deal of combinatorial
reasoning with sums and products. Thus, we had to develop some basic
theorems to support such reasoning, many of which have since been
moved into Isabelle's HOL library. These include, for example, 

\medskip

\begin{isabellebody}
\isamarkupfalse%
\isacommand{lemma}\ setsum{\isacharunderscore}cartesian{\isacharunderscore}product{\isacharcolon}\ \isanewline
\ \ \ {\isachardoublequote}{\isacharparenleft}{\isasymSum}x{\isasymin}A{\isachardot}\ {\isacharparenleft}{\isasymSum}y{\isasymin}B{\isachardot}\ f\ x\ y{\isacharparenright}{\isacharparenright}\ {\isacharequal}\ {\isacharparenleft}{\isasymSum}z{\isasymin}A\ {\isacharless}{\isacharasterisk}{\isachargreater}\ B{\isachardot}\ f\ {\isacharparenleft}fst\ z{\isacharparenright}\ {\isacharparenleft}snd\ z{\isacharparenright}{\isacharparenright}{\isachardoublequote}
\end{isabellebody}

\medskip 

\noindent which allows one to view a double summation as a sum over a 
cartesian product. A more interesting example is

\medskip

\begin{isabellebody}
\isamarkupfalse%
\isacommand{lemma}\ setsum{\isacharunderscore}reindex{\isacharcolon}\isanewline
\ \ \ \ \ {\isachardoublequote}inj{\isacharunderscore}on\ f\ B\ {\isasymLongrightarrow}\ {\isacharparenleft}{\isasymSum}x{\isasymin}f{\isacharbackquote}B{\isachardot}\ h\ x{\isacharparenright}\ {\isacharequal}\ {\isacharparenleft}{\isasymSum}x{\isasymin}B{\isachardot}\ {\isacharparenleft}h\ {\isasymcirc}\ f{\isacharparenright}{\isacharparenleft}x{\isacharparenright}{\isacharparenright}{\isachardoublequote}
\end{isabellebody}

\medskip

\noindent which expresses that if $f$ is an injective function on a 
set $B$, then summing $h$ over the image of $B$ under $f$ is
the same as summing $h \circ f$ over $B$. In particular, if $f$ is a
bijection from $B$ to $A$, the second identity implies that summing
$h$ over $A$ is the same as summing $h \circ f$ over $B$. This type of
``reindexing'' is often so transparent in mathematical arguments that
when we first came across an instance where we needed it (long ago,
when proving quadratic reciprocity), it took some thought to identify
the relevant principle. It is needed, for example, to show
\[
\sum_{d | n} h(n) = \sum_{d | n} h(n / d),
\]
using the fact that $f(d) = n / d$ is a bijection from the set of
divisors of $n$ to itself; or, for example, to show
\[
\sum_{d d' = c} h(d, d') = \sum_{d | c} h(d, c/d),
\]
using the fact that $f(d) = \la d, c /d \ra$ is a bijection from the
set of divisors of $c$ to $\{ \la d, d' \ra \st d d' = c \}$. The
reindexing lemma is a discrete analogue of integration by
substitution, so it is likely that methods developed to
support such inferences will be more generally useful.

In Isabelle, if $\sigma$ is any type, then $\fn{\sigma} \; set$ denotes
the type of all subsets of $\sigma$. The predicate ``finite'' is
defined inductively for these subset types.  Isabelle's summation
operator takes a subset $A$ of $\sigma$ and a function $f$ from
$\sigma$ to any type with an appropriate notion of addition, and
returns $\sum_{x \in A} f(x)$. This summation operator really only
makes sense when $A$ is a finite subset, so many identities have to be
restricted accordingly. (An alternative would be to define a type of
finite subsets of $\sigma$, with appropriate closure operations; but
then work would be required to translate properties of arbitrary
subsets to properties of finite subsets, or to mediate relationships
between finite subsets and arbitrary subsets.) This has the net effect
that applying an identity involving a sum or product often requires
one to verify that the relevant sets are finite. This difficulty is
ameliorated by defining $\sum_{x \in A} f(x)$ to be $0$ when $A$ is
infinite, since it then turns out that a number of identities hold in
the unrestricted form. But this fix is not universal, and so
finiteness issues tend to pop up repeatedly when one carries out a
long calculation.

In short, at present, carrying out combinatorial calculations often
requires a number of straightforward verifications involving
reindexing and finiteness. Once again, these are inferences that are
nearly transparent in ordinary mathematical texts, and so, by our
general principle, we should expect mechanized proof assistants to
take care of them. As before, there are stumbling blocks; for example,
when reindexing is needed, the appropriate injection $f$ has to be
pulled from the air. We expect, however, that in the types of
inferences that are commonly viewed as obvious, there are natural
candidates for $f$. So this is yet another domain where reflection and
empirical work should allow us to make proof assistants more usable.

\subsection{Devising elementary proofs}

Anyone who has undertaken serious work in formal mathematical
verification has faced the task of adapting an ordinary mathematical
proof so that it can be carried out using the libraries and resources
available. When a proof uses mathematical ``machinery'' that is
unavailable, one is faced with the choice of expanding the background
libraries to the point where one can take the original proof at face
value, or finding workarounds, say, by replacing the original
arguments with ones that are more elementary. The need to rewrite
proofs in such a way can be frustrating, but the task can also be
oddly enjoyable: it poses interesting puzzles, and enables one to
better understand the relationship of the advanced mathematical
methods to the elementary substitutes. As more powerful mathematical
libraries are developed, the need for elementary workarounds will
gradually fade, and with it, alas, one good reason for investing time
in such exercises.

Our decision to use Selberg's proof rather than a complex-analytic one
is an instance of this phenomenon. To this day, we do not have a sense
of how long it would have taken to build up a complex-analysis library
sufficient to formalize one of the more common proofs of the prime
number theorem, nor how much easier a formal verification of the prime
number theorem would have been in the presence of such a library.

But similar issues arose even with respect to the mild uses of
analysis required by the Selberg proof. Isabelle's real library gave
us a good theory of limits, series, derivatives, and the basic
transcendental functions, but it had almost no theory of integration
to speak of. Rather than develop such a theory, we found that we were
able to work around the mild uses of integration needed in the Selberg
proof.\footnote{Since the project began, Sebastian Skalberg managed to
  import the more extensive analysis library from the HOL theorem
  prover to Isabelle. By the time that happened though, we had already
  worked around most of the applications of analysis needed for the
  proof.} Often, we also had to search for quick patches to other gaps
in the underlying library. For the reader's edification and
entertainment, we describe a few such workarounds here.

Recall that one of the fundamental identities we needed asserts
\[
\ln(1 + 1/n) = 1/n + O(1/n^2).
\] 
This follows from the fact that $\ln(1 + x)$ is well approximated by
$x$ when $x$ is small, which, in turn, can be seen from the
Maclaurin series for $\ln(1 + x)$, or even the fact that the
derivative of $\ln(1 + x)$ is equal to $1$ at $0$. But these were
among the few elementary properties of transcendental functions that
were missing from the real library. How could we work around this?

To be more specific: Fleuriot's real library defined $e^x$ by the
power series $e^x = \sum_{n = 0}^{\infty} x^n / n!$, and showed that
$e^x$ is strictly increasing, $e^0 = 1$, $e^{x + y} = e^x e^y$ for
every $x$ and $y$, and the range of $e^x$ is exactly the set of
positive reals. The library then defines $\ln$ to be a left inverse to
$e^x$. The puzzle was to use these facts to show that $| \ln(1 + x) -
x| \leq x^2$ when $x$ is positive and small enough.

Here is the solution we hit upon. First, note that when $x \geq 0$,
$e^x \geq 1 + x$, and so, $x \geq \ln(1 + x)$. Replacing $x$ by $x^2$,
we also have 
\begin{equation}
\label{e:eq:one}
e^{x^2} \geq 1 + x^2. 
\end{equation}
On the other hand, the definition of $e^x$ can be used to show
\begin{equation}
\label{e:eq:two} 
e^x \leq 1 + x + x^2
\end{equation}
when $0 \leq x \leq 1/2$. From (\ref{e:eq:one}) and (\ref{e:eq:two})
we have
\[
\begin{split}
e^{x - x^2} & = e^x / e^{x^2} \\
& \leq (1 + x + x^2) / (1 + x^2) \\
& \leq 1 + x,
\end{split}
\]
where the last inequality is easily obtained by multiplying
through. Taking logarithms of both sides, we have
\[
x - x^2 \leq \ln(1 + x) \leq x
\]
when $0 \leq x \leq 1/2$, as required. In fact, a similar calculation
yields bounds on $\ln(1 + x)$ when $x$ is negative and close to
$0$. This can be used to show that the derivative of $\ln x$ is $1 /
x$; the details are left to the reader.

For another example, consider the problem of showing that $\sum_{n =
  1}^\infty 1 / n^2$ converges. This follows immediately from the
integral test: $\sum_{n = 1}^\infty 1 / n^2 \leq \int_1^\infty 1 / x^2
= 1$. How can it be obtained otherwise? Answer: simply write
\begin{eqnarray*}
\sum_{n = 1}^M 1 / n^2 & \leq & 1 + \sum_{n = 2}^M 1 / n (n - 1) \\ 
& = & 1 + \sum_{n = 2}^M (1 / (n -1 ) - 1 / n) \\
& = & 1 + 1 - 1/M \\
& \leq & 2,
\end{eqnarray*}
where the second equality relies on the fact that the preceding
expression involves a telescoping sum. Having to stop frequently to
work out puzzles like these helped us appreciate the immense power of
the Newton-Leibniz calculus, which provides uniform and mechanical
methods for solving such problems. The reader may wish to consider
what can be done to show that the sum $\sum_{n = 1}^\infty 1 / x^a$ is
convergent for general values of $a > 1$, or even for the special case
$a = 3 / 2$. Fortunately, we did not need these facts.

Now consider the identity 
\[
\sum_{n \leq x} 1 / n = \ln x +
O(1). 
\]
To obtain this, note that when $x$ is positive integer we can write
$\ln x$ as a telescoping sum,
\[
\begin{split}
\ln x & = \sum_{n \leq x - 1} (\ln (n + 1) - \ln n) \\
& = \sum_{n \leq x - 1} \ln(1 + 1 / n) \\
& = \sum_{n \leq x - 1} 1 / n + O(\sum_{n \leq x} 1 / n^2) \\
& = \sum_{n \leq x} 1 / n + O(1).
\end{split}
\]
We learned this trick from Cornaros and Dimitracopoulos
\cite{cornaros:dimitracopoulos:94}. In fact, a slight refinement of
the argument shows
\[
\sum_{n \leq x} 1 /n = \ln x + C + O(1/x)
\]
for some constant, $C$. This constant is commonly known as Euler's
constant, denoted by $\gamma$. 

One last puzzle: how can one show that $\ln x / x^a$ approaches $0$,
for any $a > 0$? Here is our solution. First, note that we have $\ln x
\leq \ln(1 + x) \leq x$ for every positive $x$. Thus we have
\[
a \ln x = \ln x^a \leq x^a,
\]
for every positive $x$ and $a$. Replacing $a$ by $a / 2$ and dividing
both sides by $a x^a / 2$, we obtain $\ln x / x^a \leq 2 /
(ax^{a/2})$. It is then easy to show that the right-hand-side
approaches $0$ as $x$ approaches infinity.

\section{Conclusions}
\label{conclusions:section}

Our efforts show that formal verification of significant mathematical
theorems is possible, although more work is needed before the practice
is likely to become widespread. In an ideal situation, it would be
possible to enter mathematical text almost exactly as it appears in a
careful and precise informal presentation, and interactive proof
systems would be able to verify inferences at that level. Our
formalization of the prime number theorem provides a case study that
clarifies some of the ways in which the current technology
falls short of the ideal.

The formal statements of theorems in
Section~\ref{formalization:subsection} are notably less attractice
than their informal counterparts in Section~\ref{selberg:subsection}.
The difference is not merely cosmetic; notation is an integral part of
mathematics, and it is unreasonable to expect the mathematical
community to make notational sacrafices for mechanical convenience.
Integrating formal verification into mathematical practice will
therefore require us to take ordinary mathematical notation extremely
seriously.

The biggest obstacle at present is the gap between those inferences
that ordinary mathematicians recognize as obvious, and those that can
be verified automatically by conventional proof assistants. We have
suggested one strategy for improvement, namely, to reflect on the
capacities that enable us, in specific domains, to verify textbook
inferences, and then to formalize that understanding. In particular,
it seems that fairly straightforward support for reasoning about
inequalities between real numbers and casts between integers and real
numbers would have simplified our task substantially.

Progress in formal verification will require a broad but focused
philosophical reflection on ordinary mathematical practice, together
with robust formal characterizations of that practice and sound
engineering. As such, the field represents an auspicious combination
of theory and practice.

\bigskip

\noindent {\bf Acknowledgements.} We are especially grateful to Tobias
Nipkow and Larry Paulson for continued support on this project. We are
also grateful to them, and to Freek Wiedijk and two anonymous
referees, for comments on this paper.

\bibliographystyle{plain} 
\bibliography{proofthry,numbertheory,automated}

\end{document}